\documentclass[runningheads]{llncs}

\usepackage{eccv}

\usepackage{eccvabbrv}
\usepackage{wrapfig}
\usepackage{multirow}
\usepackage{graphicx}
\usepackage{booktabs}
\usepackage{wrapfig}
\usepackage[accsupp]{axessibility}  
\usepackage{pifont}%
\newcommand{\cmark}{{\color{green} \ding{51}}}%
\newcommand{\xmark}{{\color{red} \ding{55}}}%

\usepackage{todonotes}

\usepackage{hyperref}

\usepackage{orcidlink}

\begin{document}

\title{FreeTalk: Emotional Topology-Free \\ 3D Talking Heads} 


\author{Federico Nocentini$^*$\inst{1}
\and Thomas Besnier$^*$\inst{2}
\and  Claudio Ferrari\inst{1}
\and  Stefano Berretti\inst{1}
\and Mohamed Daoudi\inst{3,4}}
\authorrunning{F. Nocentini et al.}
%

\institute{
 Media Integration and Communication Center (MICC),\\ University of Florence, Italy\\ 
 \email{federico.nocentini@unifi.it, stefano.berretti@unifi.it, claudio.ferrari@unifi.it}
\and University of Copenhagen
 \email{thomas.besnier@di.ku.dk} \\ \and
IMT Nord Europe, Institut Mines-Télécom, Centre for Digital Systems
\and
Univ. Lille, CNRS, Centrale Lille, Institut Mines-Telecom, UMR 9189 CRIStAL, F-59000 Lille, France\\
\email{mohamed.daoudi@imt-nord-europe.fr}\\
}

\maketitle
\def\thefootnote{*}\footnotetext{Equal contribution}

\begin{abstract}
\vspace{-0.5cm}
Speech-driven 3D facial animation has advanced rapidly, yet most approaches remain tied to registered template meshes, preventing effective deployment on raw 3D scans with arbitrary topology. At the same time, modeling controllable emotional dynamics beyond lip articulation remains challenging, and is often tied to template-based parameterizations. 
We address these challenges by proposing \textbf{FreeTalk}, a two-stage framework for \emph{emotion-conditioned} 3D talking-head animation that generalizes to \emph{unregistered} face meshes with arbitrary vertex count and connectivity. 
First, \textbf{Audio-To-Sparse (ATS)} predicts a temporally coherent sequence of 3D landmark displacements from speech audio, conditioned on an emotion category and intensity. This sparse representation captures both articulatory and affective motion while remaining independent of mesh topology. 
Second, \textbf{Sparse-To-Mesh (STM)} transfers the predicted landmark motion to a target mesh by combining intrinsic surface features with landmark-to-vertex conditioning, producing dense per-vertex deformations without template fitting or correspondence supervision at test time. 
Extensive experiments show that FreeTalk matches specialized baselines when trained in-domain, while providing substantially improved robustness to unseen identities and mesh topologies. Code and pre-trained models will be made publicly available.
\keywords{3D Talking Heads \and Topology-agnostic deformation \and Emotion-conditioned  \and Landmark-based motion \and Intrinsic surface learning}
\end{abstract}

\begin{figure}
    \centering
    \includegraphics[width=0.68\linewidth]{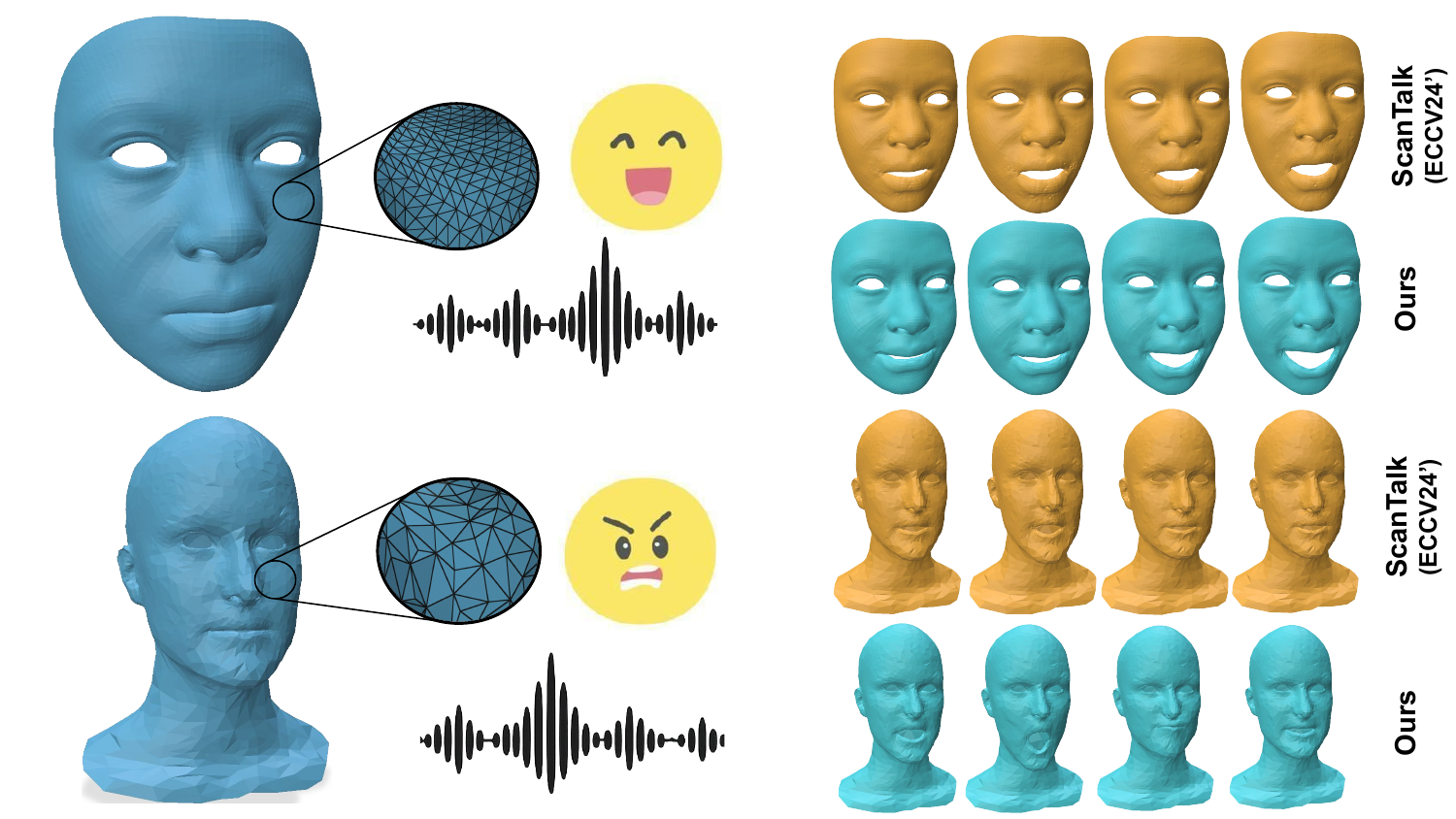}
    \vspace{-0.3cm}
    \caption{\textbf{FreeTalk} takes an unregistered static face mesh (in \textcolor{blue}{blue} on the left) and an emotional audio speech and predicts a sequence of deformation fields (in \textcolor{cyan}{cyan}) matching speech-related motions and expression-related deformations.}
    \label{fig:thumbnail}
\end{figure}

\section{Introduction}
\label{sec:intro}

Human face animation is a long-standing and challenging task in Computer Vision, with applications spanning virtual reality, digital humans, movie production, and video games. In recent years, speech-driven facial animation has emerged as a key, application-oriented research direction, aiming to synthesize temporally coherent and visually realistic 3D facial motion directly from audio signals.
Despite significant progress, learning a robust cross-modal mapping from speech to 3D facial geometry remains fundamentally ill-posed. Audio alone provides incomplete information about facial motion: identical phonemes may correspond to different visual realizations depending on speaker identity, co-articulation effects, and, crucially, emotional state. Emotional content introduces complex, coordinated deformations that extend beyond lip articulation, involving eyebrows, cheeks, and global facial musculature. Modeling such affective dynamics jointly with speech remains a challenging problem.
An additional and often overlooked limitation of current methods lies in their dependency on a fixed arrangement of triangles (number of vertices/faces and face connectivity) of the input mesh: the \textbf{mesh topology} is predetermined. Most state-of-the-art approaches operate on registered template meshes (e.g., FLAME-based \cite{FLAME:SiggraphAsia2017} models), implicitly binding the learned representation to a specific vertex connectivity. This design choice simplifies learning but severely limits generalization: newly acquired 3D scans must be fitted to the template before animation can be applied, introducing computational overhead and reducing practical applicability.
%
Existing works partially address either emotional expressiveness~\cite{emote, emoface, peng2023emotalk} or topology generalization~\cite{nocentini2024scantalk3dtalkingheads}, but not both simultaneously. Several approaches incorporate emotion control, either via explicit emotion labels~\cite{nocentini2025emovocaspeechdrivenemotional3d} or implicit audio embeddings, yet they operate on fixed, registered meshes. 

In this work, as described in~\Cref{fig:thumbnail}, we bridge this gap by proposing the first framework that jointly models emotionally expressive speech-driven facial motion and enables animation across arbitrary mesh topologies without template registration.
Our approach is based on a two-stage architecture that explicitly decouples motion generation from geometric realization. In the first stage, we train an identity-agnostic generative model that takes as input speech audio together with emotion category and intensity labels, and predicts a sequence of 3D landmark displacements. This intermediate representation captures both articulatory and affective facial dynamics in a compact, topology-independent form.
In the second stage, we learn a geometry-aware deformation module that transfers the predicted landmark motion onto any target 3D face mesh. Given the mesh geometry and the landmark displacements, the model predicts consistent vertex deformations, enabling animation of arbitrary topologies without requiring registration to a template. By explicitly separating expressive motion synthesis from identity and topology-specific deformation, our framework achieves both emotional controllability and topology invariance. This decoupled design not only improves generalization to unseen identities and meshes, but also removes the need for costly template fitting procedures, facilitating practical deployment.

Our contributions are three-fold. First, we introduce \textbf{Audio-To-Sparse} (ATS), an identity-agnostic generative model designed to produce emotionally expressive landmark displacement sequences conditioned on speech, emotion category, and intensity. Building upon this, we propose \textbf{Sparse-To-Mesh} (STM), a discretization-agnostic deformation module that transfers these sparse motions to arbitrary 3D face meshes without the need for fixed connectivity or template registration. Finally, through qualitative and quantitative evaluation, we support that FreeTalk successfully combines expressive speech-driven animation with topology generalization, addressing a significant limitation of prior works.


    

\section{Related Work}
\label{sec:related}

3D facial animation methods are commonly built by registering raw scans \cite{bosphorus_2008, Texas3D_2010} to a predefined template topology~\cite{FLAME:SiggraphAsia2017, COMA:ECCV18, muralikrishnan_2023_BLISS, wuu2022multiface, BIWI_2010, Li_2020_Dynamic_facial_asset_and_rig_generation_from_a_single_scan} and subsequently animating them using 3D morphable models (3DMMs)~\cite{morphable_models_review}. While effective, this registration step may smooth fine geometric details and tightly couples animation to a fixed connectivity.
Several learning-based approaches have proposed latent representations of raw face scans using point-based encoders such as PointNet~\cite{Liu_2019, Bahri_SMF_2021, Croquet_Diff_Reg_OT_2021, Charles_PointNet_2017, varifold_loss, shape_transformer, qi2017pointnet++}. 
For robust surface feature learning, models such as DiffusionNet~\cite{SharpDiffusionNet} or PoissonNet~\cite{maesumi2025poissonnet} can be combined with robust deformation methods such as jacobian fields~\cite{NeuralJacobianField_2022}.
While promising, these approaches are general method, not yet specialized for speech-driven animation.
\vspace{-3mm}
\paragraph{Speech-Driven 3D Facial Animation.}
Early facial animation systems relied on procedural rigs and rule-based viseme blending~\cite{DEMARTINO2006971,10.1145/2897824.2925984,982373}, manually mapping phonemes to facial controls. The availability of large-scale 3D capture enabled data-driven modeling. Karras~\etal~\cite{10.1145/3072959.3073658} demonstrated high-fidelity actor-specific reconstruction from dense 3D capture, though without cross-identity generalization. VOCA~\cite{VOCA2019} extended speech-driven animation to multi-subject settings.
Subsequent works adopted stronger generative architectures. \textit{MeshTalk}~\cite{richard2021meshtalk} introduced categorical autoregressive latent modeling, while \textit{FaceFormer}~\cite{fan2022faceformer} leveraged Transformers~\cite{NIPS2017_attention_is_all_you_need} to capture long-range temporal dependencies and more recent approaches~\cite{xing2023codetalker, peng2023selftalk, UniTalker_2024, KMTalk_2024} improved the accuracy and lip sync. Diffusion-based approaches, such as \textit{FaceDiffuser}~\cite{FaceDiffuser_Stan_MIG2023} and DiffPoseTalk~\cite{sun2024diffposetalkspeechdrivenstylistic3d}, reformulated speech-driven animation as a denoising process. Personalized motion control was explored in \textit{Imitator}~\cite{Thambiraja_2023_ICCV}. 
Despite architectural advances, these methods typically assume registered template meshes and operate on neutral speech datasets such as VOCASET or custom datasets.

\vspace{-3mm}
\paragraph{Emotionally Expressive Talking Heads.}
To model affective dynamics, several works introduced emotion-aware speech-driven animation. Reconstruction-based pipelines regress expressive parameters from 2D video using models such as EMOCA~\cite{danvevcek2022emoca}, generating pseudo-3D expressive supervision~\cite{lu2023audiodriven}. Other approaches inject emotional cues from audio embeddings, as in \textit{EmoTalk}~\cite{peng2023emotalk}, or target specific expressive behaviors such as laughter in \textit{LaughTalk}~\cite{sungbin2023laughtalk}. 
Datasets such as EMOTE~\cite{emote} and EmoVOCA~\cite{nocentini2025emovocaspeechdrivenemotional3d} enabled explicit emotion supervision in 3D space. More recent systems such as EmoFace~\cite{emoface}, DEEPTalk~\cite{deeptalk_2025}, MEDTalk~\cite{medtalk}, and MemoryTalker~\cite{MemoryTalker_2025} explore dynamic emotion control, probabilistic expressive motion, and personalized stylization. These methods improve realism but are still bound to registered meshes or predefined rig spaces.
\vspace{-3mm}
\paragraph{Topology-Agnostic Facial Animation.}
To alleviate the fixed mesh topology limitation, intermediate geometry-aware representations have been explored such as Neural Face Rigging~\cite{Qin_2023_NFR} and Neural Face Skinning~\cite{cha2025neural_face_skinning} for face motion retargeting but these approaches are not speech-driven. Building on this idea, \textit{ScanTalk}~\cite{nocentini2024scantalk3dtalkingheads} introduced the first topology-agnostic framework for speech-driven 3D talking head animation, learning a connectivity-independent motion representation transferable to arbitrary meshes. While successfully removing the fixed-topology constraint, ScanTalk focuses on neutral speech and does not explicitly model emotional dynamics.
In contrast, FreeTalk jointly addresses affective modeling and mesh generalization, enabling animation of arbitrary unregistered face meshes directly from audio while preserving geometric fidelity and lip synchronization quality in a multi-modal 4D setting.

\begin{wraptable}{r}{0.48\linewidth} 
\centering
\vspace{-0.6cm}
\resizebox{\linewidth}{!}{%
\begin{tabular}{lcc}
\hline
        & Emotion & Any mesh \\ \hline
CodeTalker / FaceFormer / SelfTalk  &  \xmark   &   \xmark  \\
EmoTalk / Emote / Emovoca &  \cmark   &   \xmark   \\
ScanTalk  &  \xmark   &    \cmark                \\ \hline
\textbf{FreeTalk}    &     \cmark       &    \cmark \\
\hline
\end{tabular}
}
\vspace{-0.2cm}
\caption{Comparison of talking head models.}
\vspace{-0.6cm}
\label{tab:limitation_comparison}
\end{wraptable}

In summary, as shown in~\Cref{tab:limitation_comparison}, existing works address either emotionally expressive speech-driven animation under fixed mesh topologies, or topology-agnostic animation under neutral conditions. To the best of our knowledge, no prior method jointly models affective speech-driven facial motion while generalizing across arbitrary topologies without template registration. Our work fills this gap by decoupling emotion-aware motion generation from topology-specific geometric deformation.






\vspace{-0.3cm}
\section{Proposed Approach: FreeTalk}
\begin{figure}[!th]
    \centering
    \includegraphics[width=\linewidth]{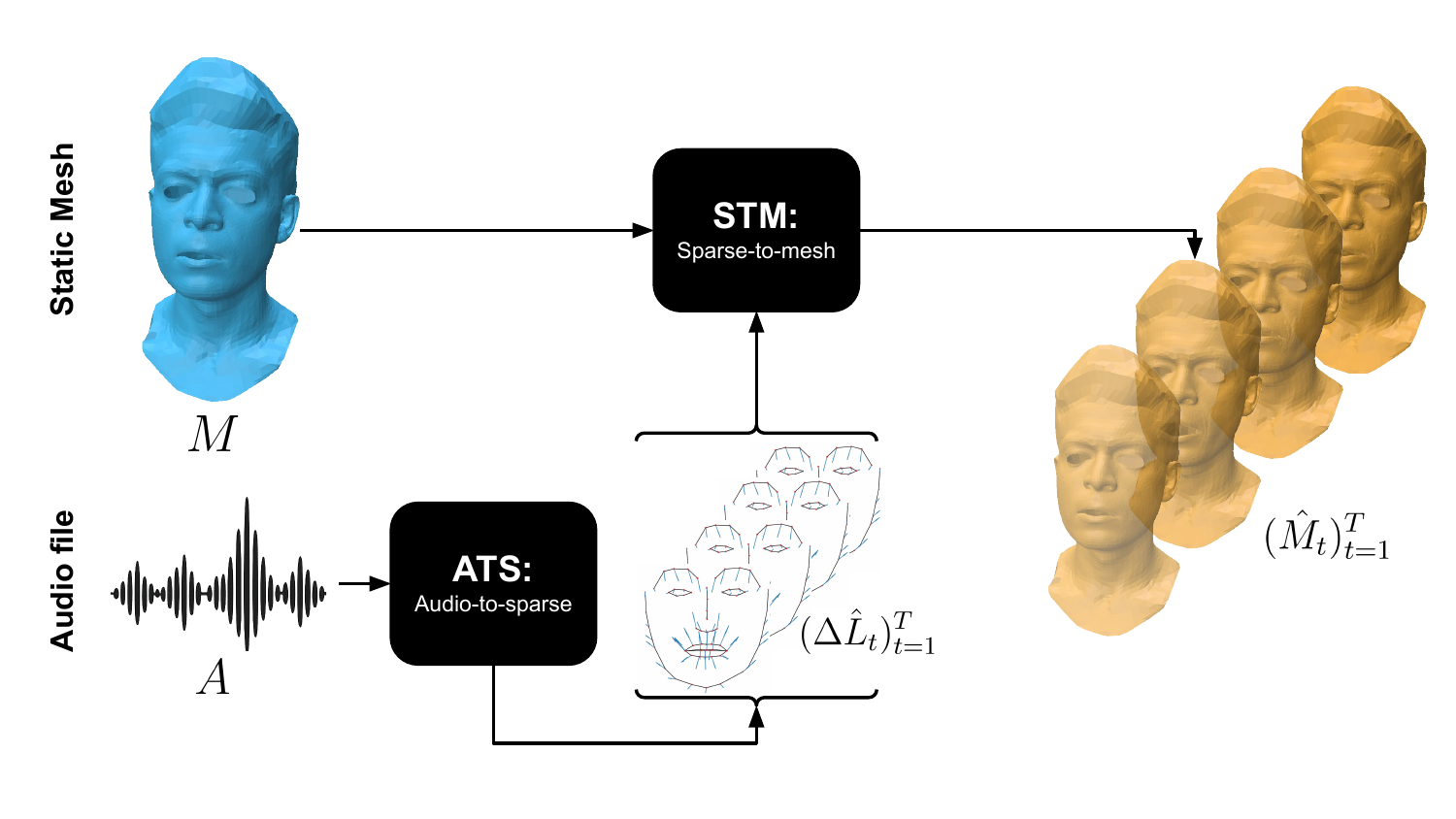}
    \caption{\textbf{Overview of the proposed approach.} ATS takes the speech audio signal $A$ and generates a sequence of landmarks displacements $(\Delta L_t)_{t=1}^{T} $. Then, STM takes a static neutral mesh $M$ and maps the sparse motion sequence of landmarks $(\Delta L_t)_{t=1}^{T} $ to dense vertex motions to predict the final animated mesh sequence $(\hat{M}_t)_{t=1}^{T} $.}
    \label{fig:full_overview}
    \vspace{-0.5cm}
\end{figure}

We propose \textbf{FreeTalk} a two-stage framework, as shown in~\Cref{fig:full_overview},  for speech-driven 3D facial animation that generalizes across \emph{arbitrary} mesh topologies by explicitly decoupling expressive speech motion synthesis from geometric realization. Given a target face mesh $M=(V,F)$, where $V\in\mathbb{R}^{n\times 3}$ denotes vertex positions and $F\in\mathbb{N}^{m\times 3}$ the mesh connectivity, and an input speech signal $A$ of duration $T$, FreeTalk generates a temporally coherent sequence of deformed meshes $(\hat{M}_t)_{t=1}^{T}$ without requiring template registration, fixed vertex ordering, or any constraint on topology. The core idea is to first generate identity-agnostic facial motion in a sparse, topology-independent representation, and then transfer this motion onto the specific target geometry. Concretely, FreeTalk comprises (i) an \textbf{Audio-To-Sparse (ATS)} module that predicts a sequence of 3D landmark displacements conditioned on speech, emotion category and emotion intensity, and (ii) a \textbf{Sparse-To-Mesh (STM)} module that maps the predicted landmark displacements motion to a dense vertex displacement field over the target mesh. 
This decomposition stems from the observation that the facial musculoskeletal structure induces neighboring points to move according to consistent motion patterns~\cite{ferrari2021sparse}, making it possible to exploit such correlation to learn a dense motion field from a sparse set of points. This further allows separating expressive motion generation from mesh-specific deformation and animate arbitrary face meshes while preserving geometric consistency. This strategy has been explored in previous works for 4D face animation yet on fixed topology meshes~\cite{otberdout2022sparse, nocentini2023learning,otberdout2023generating}. We here generalize it to work on arbitrary mesh structures. 

\subsection{ATS: Audio-To-Sparse}
\label{sec:ats}
The goal of the \textbf{Audio-To-Sparse (ATS)} module is to generate a temporally coherent 3D facial motion sequence directly from speech. Rather than predicting dense geometry, ATS operates in a sparse landmark space, which provides a compact yet expressive representation of facial dynamics and serves as an intermediate abstraction between audio and full mesh animation.
Let $T$ represent the total number of animation frames corresponding to the input speech signal. For each frame $t \in \{1,\dots,T\}$, the facial configuration is represented by $N=68$ landmarks $L_t \in \mathbb{R}^{N \times 3}$,
which capture semantically meaningful facial keypoints (e.g., lips, jaw, eyebrows). These landmarks provide sufficient geometric structure to capture articulation and expression while remaining lightweight. Then, to decouple identity from motion, we assume a subject-specific neutral template with $L^{\mathrm{tpl}} \in \mathbb{R}^{N \times 3}$
and represent facial motions as a curve, parametrized by $t$ in a space of landmark displacements:
\begin{equation}
\Delta L_t = L_t - L^{\mathrm{tpl}}.
\end{equation}
This formulation allows the network to focus on motion dynamics while preserving the identity. By stacking the landmark displacements into a vector of shape $D=3\times N$ the complete motion sequence can be compactly written as:
\begin{equation}
x_0 = (\Delta L_t )_{t=1}^{T} \in \mathbb{R}^{T \times D}.
\end{equation}
ATS is therefore formulated as learning a conditional generative mapping
\begin{equation}
\mathrm{ATS}: (A, e, i) \longmapsto x_0,
\end{equation}
where $e$ and $i$ denote discrete emotion category and intensity label, respectively. This explicit conditioning enables controllable affect synthesis while maintaining speech-driven articulation.
\vspace{-3mm}
\paragraph{Audio representation.}
To extract high-level speech features, we use a pretrained HuBERT encoder~\cite{Hubert_audio_encoding}. 
Given a speech signal $A$, the encoder produces contextualized token-level representations
$H = \mathrm{HuBERT}(A) \in \mathbb{R}^{S \times C} (C=768)$,
where $S$ depends on the internal temporal resolution of the encoder. 
These features are then resampled to match the landmark timeline, yielding
$\tilde{H} \in \mathbb{R}^{T \times C}$.
The sequence $\tilde{H}$ provides one speech embedding per animation frame and encodes phonetic content, and longer-range temporal context. This representation serves as the conditioning memory for motion generation.
\vspace{-3mm}
\paragraph{Conditional diffusion model.}
We model the conditional distribution $p(x_0 \mid A, e, i)$ using a denoising diffusion probabilistic model~\cite{DDPM_2020} defined over full motion trajectories. Diffusion models are particularly suited to this setting as they allow stable training on high-dimensional structured outputs and naturally capture multi-modal distributions.
Let $\{\beta_\ell\}_{\ell=1}^{T_d}$ be a linear noise schedule with $T_d$ diffusion steps. Define $\alpha_\ell = 1 - \beta_\ell$ and $\bar{\alpha}_\ell = \prod_{j=1}^{\ell} \alpha_j$. The forward diffusion process gradually corrupts the clean sequence $x_0$ to generate the noisy sequence $x_\ell$ as:
\begin{equation}
q(x_\ell \mid x_0)
=
\mathcal{N}\!\left(
x_\ell;
\sqrt{\bar{\alpha}_\ell}\,x_0,
(1-\bar{\alpha}_\ell)\mathbf{I}
\right),
\end{equation}
which can be sampled in closed form as:
\begin{equation}
x_\ell
=
\sqrt{\bar{\alpha}_\ell}\,x_0
+
\sqrt{1-\bar{\alpha}_\ell}\,\epsilon,
\qquad
\epsilon \sim \mathcal{N}(0,\mathbf{I}).
\label{eq:ats_forward}
\end{equation}
A conditional denoiser $f_\theta$ is trained to reconstruct $\hat{x}_0 = f_\theta(x_\ell, \ell, A, e, i)$, 
where $\ell$ is sampled uniformly during training and $\hat{x}_0$ is the predicted landmarks displacements sequence. Intuitively, the network is asked to reconstruct the clean motion sequence from progressively noisier versions while being guided by speech and affect cues. This formulation enables the model to learn a rich mapping from audio dynamics to facial motion trajectories.

\vspace{-3mm}
\paragraph{Transformer-based conditional denoiser.}
The denoiser $f_\theta$ is implemented as a Transformer decoder operating on the noisy landmark sequence with cross-attention to audio features. 
The noisy displacement sequence $x_\ell \in \mathbb{R}^{T \times D}$ and the audio $\tilde{H} \in \mathbb{R}^{T \times C}$ features are projected to the model dimension $d$ and augmented with learned positional embeddings:
\begin{equation}
X_\ell = W_x x_\ell + P_{\mathrm{tgt}} \in \mathbb{R}^{T \times d}, \qquad \hat{A} = W_h \tilde{H} + P_{\mathrm{mem}} \in \mathbb{R}^{T \times d} .
\end{equation}
Here $W_h$ and $W_x$ are learned linear projections and $P_{\mathrm{mem}}$ and $P_{\mathrm{tgt}}$ denote learned positional encodings.
Conditioning information $X_\ell$ is injected into the target tokens through (i) a sinusoidal embedding of the diffusion timestep $\ell$ followed by a Multi-Layer Perceptron (MLP) $g_t(\ell)$, and (ii) an affect embedding $g_c(e,i)$. The resulting decoder input tokens are given by $Z = X_\ell + g_t(\ell) + g_c(e,i)$.
Each decoder layer first applies multi-head self-attention $Z' = \mathrm{SelfAttn}(Z)$ allowing temporal interactions across landmark tokens.
Then, cross-attention is performed between the updated landmark tokens $Z'$ and the audio memory $\hat{A}$. Denoting by $Q$, $K$, and $V$ the projected queries, keys, and values, we have:
\begin{equation}
Q = Z' W_Q, \qquad
K = \hat{A} W_K, \qquad
V = \hat{A} W_V.
\end{equation}
The cross-attention update is then given by:
\begin{equation}
\mathrm{Attn}(Z', \hat{A})
=
\mathrm{softmax}\!\left(
\frac{QK^\top}{\sqrt{d}}
\right)V.
\end{equation}
Through this hierarchical attention mechanism, the model first refines the noisy landmark trajectory via global temporal interactions and then aligns it with the speech representations. This design enables the denoiser to jointly capture intrinsic facial motion coherence and audio-conditioned articulatory and affective dynamics at each diffusion step.

\vspace{-3mm}
\paragraph{Monotonic cross-attention constraint.}
To further emphasize temporally consistent behavior, we introduce a band-diagonal constraint on the cross-attention weights. Specifically, attention from frame $i$ to audio token $j$ is allowed only within a fixed radius $r$:
\begin{equation}
B_{ij} =
\begin{cases}
0 & \text{if } |j-i| \le r, \\
1 & \text{otherwise}.
\end{cases}
\end{equation}
The masked cross-attention then becomes:
\begin{equation}
\mathrm{Attn}_{\mathrm{mono}}(Z', \hat{A})=
\mathrm{softmax}\!\left(\frac{QK^\top}{\sqrt{d}} + B\right)V.
\end{equation}
This inductive bias promotes an approximately-monotonic correspondence between audio progression and facial motion yet allowing contextual interaction.

\vspace{-2mm}
\paragraph{Prediction head.}
Let $Z_{\mathrm{out}} \in \mathbb{R}^{T \times d}$ be the output tokens of the final Transformer decoder layer. 
We map them back to the landmarks displacement space with a linear projection:
\begin{equation}
\hat{x}_0 = Z_{\mathrm{out}} W_o + b_o,
\qquad
W_o \in \mathbb{R}^{d \times D},\; b_o \in \mathbb{R}^{D}.
\end{equation}
Reshaping each frame $\hat{x}_{0,t} \in \mathbb{R}^{D}$ into $\mathbb{R}^{N \times 3}$ yields the predicted landmark displacements $(\hat{\Delta L}_t)_{t=1}^{T}$.


\vspace{-3mm}
\paragraph{Sampling and output representation.}
At inference time, motion sequences are generated by simulating the learned reverse diffusion process conditioned on speech and affect. Starting from Gaussian noise $x_{T_d} \sim \mathcal{N}(0, \mathbf{I})$,
we iteratively apply the conditional denoiser $f_\theta$ using a DDIM sampler over $N$ reverse steps. At each step $\ell \rightarrow \ell-1$, the model predicts an estimate of the clean displacement sequence $\hat{x}_0 = f_\theta(x_\ell, \ell, A, e, i)$, which is used to update the latent trajectory according to the deterministic DDIM dynamics.
After the final reverse step, we obtain the predicted displacement sequence $\hat{x}_0 \in \mathbb{R}^{T \times D}$.
Each frame $\hat{x}_{0,t} \in \mathbb{R}^{D}$ is reshaped into $\hat{\Delta L}_t \in \mathbb{R}^{N \times 3}$, yielding the landmark displacement trajectory $(\hat{\Delta L}_t)_{t=1}^{T}$. 
This sequence represents frame-wise landmark displacements relative to the neutral template and constitutes the output of the ATS module.
ATS is the first stage of our multi-module pipeline: the predicted displacement sequence $\hat{x}_0$ is then fed into the subsequent module, which is explicitly trained to operate in the displacement space. Working with this representation preserves identity–motion disentanglement, since the neutral template $L^{\mathrm{tpl}}$ remains factored out and static across the sequence. 

\vspace{-3mm}
\subsection{STM: Sparse-To-Mesh}

The Sparse-To-Mesh module converts the topology and identity-independent landmark motion predicted by ATS into a sequence of dense per-vertex deformation fields. STM takes a static mesh $M=(V, F)$ in neutral state with vertices $V \in \mathbb{R}^{n \times 3}$ and faces $F$, and a sequence of predicted landmark displacements $(\Delta \hat{L}_t)_{t=1}^{T}$ (from ATS) where $\Delta \hat{L}_t \in \mathbb{R}^{N \times 3}$. For each time step $t$, STM predicts a sequence of vertex displacement fields $(\Delta V_t)_{t=1}^{T}$ with $\Delta V_t \in \mathbb{R}^{n \times 3}$ on $M$. The deformed mesh at frame $t$ is then obtained as $\hat{M}_t = (\hat{V}_t, F)$ where $\hat{V}_t = V + \Delta V_t$.

\begin{figure}[!th]
    \centering
    \includegraphics[width=\linewidth]{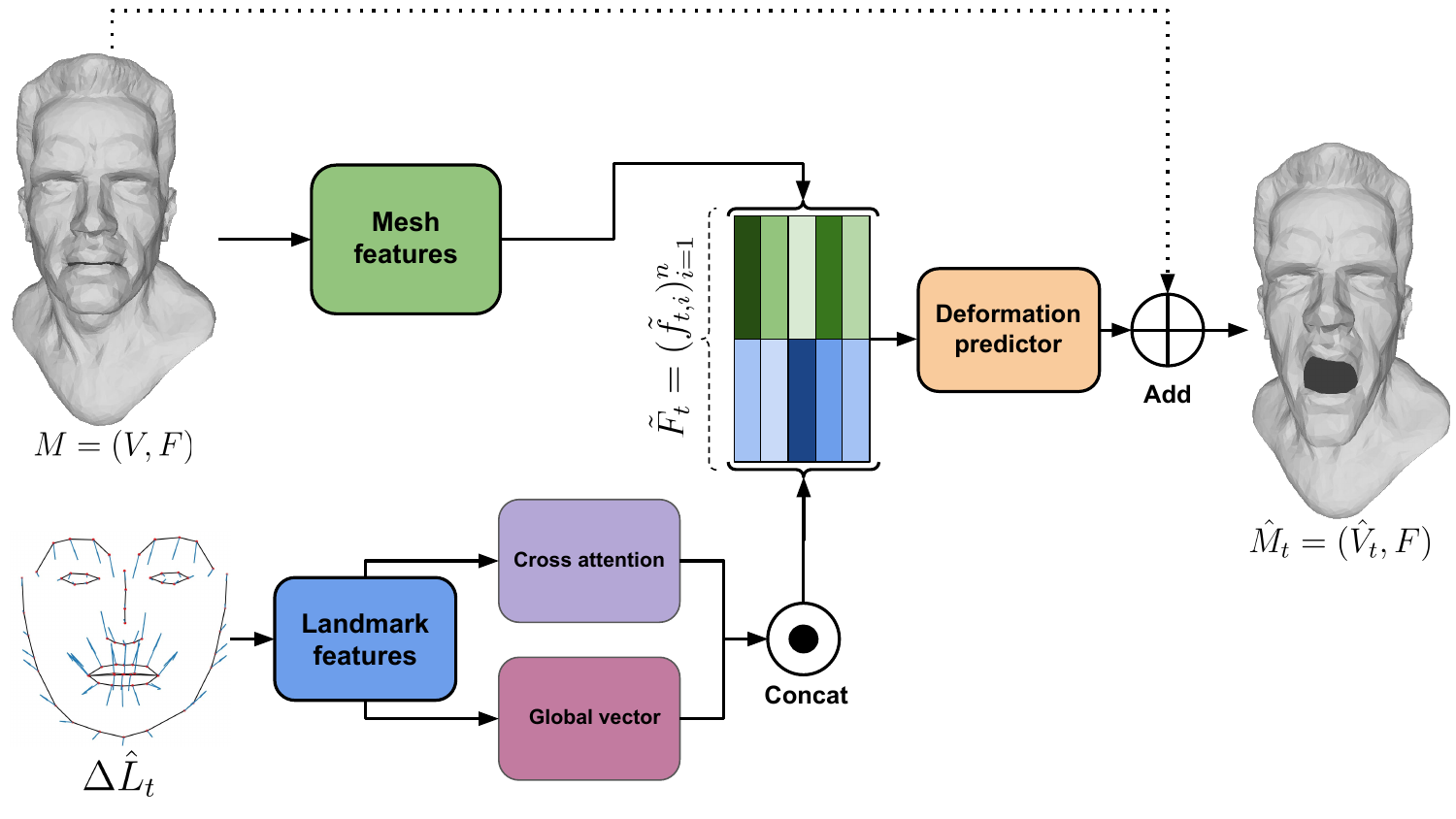}
    \caption{\textbf{Overview of the Sparse-To-Mesh (STM) module.} STM predicts vertex-wise features on a static mesh $M$, and landmark features from displacement vectors $L$. The mesh features are enhanced by a learnable mapping that injects a global embedding of the landmark displacements into each vertex via cross-attention.}
    \label{fig:LMK_to_MESH}
    \vspace{-5mm}
\end{figure}

A key requirement is that STM must generalize across arbitrary mesh triangulations, without relying on prior registration. This motivates the use of intrinsic, discretization-agnostic surface learning for mesh features. As illustrated in~\Cref{fig:LMK_to_MESH}, STM consists of four stages. First, static mesh features are extracted once from the input neutral mesh and shared across all time steps. Second, for each frame, sparse landmark motion features are computed from the predicted displacement vectors. Third, a learned soft correspondence mechanism injects motion information into the mesh representation by combining cross-attention between landmark and mesh features with a global displacement embedding that is broadcast to all vertices. Finally, the fused per-vertex features are decoded into dense vertex displacement fields, producing frame-wise mesh deformations.
\vspace{-2mm}
\paragraph{Static mesh feature extractor (dense).}
We first compute vertex-wise features on the static input mesh $M$ using a mesh encoder $E_M$. For each vertex $i$, the encoder predicts a feature vector $f_i \in \mathbb{R}^{d}$ so that $E_M(M) = (f_i)_{i=1}^{n}$.
To make the model robust to remeshing and changes in discretization, we use DiffusionNet~\cite{SharpDiffusionNet}, which is intrinsic to the mesh geometry. The encoder takes as input vertex coordinates and associated normal vectors. Since the input mesh is static, the dense feature field is computed once and reused for all frames.

\vspace{-2mm}
\paragraph{Landmark motion encoder (sparse).}
For each frame $t$, the landmark displacement set $\Delta \hat{L}_t \in \mathbb{R}^{N \times 3}$ is processed in two complementary ways. First, all landmark displacements are stacked into a single global motion vector $\Delta \hat{L}_t \in \mathbb{R}^{D}$, 
which is later replicated across all vertices and provides global motion context. Second, we compute landmark-wise features using a lightweight Graph Convolutional Network (GCN) defined on the fixed landmark graph $G_L = (\mathcal{N}_L, \mathcal{E}_L)$. For each landmark $j$, the GCN predicts $l_{t,j} = \mathrm{GCN}(\Delta \hat{L}_t)_j$.

Because two landmarks may exhibit similar displacements within the same frame, we augment the landmark features with a positional encoding based on its index as
$p_j = \mathrm{PE}(j)$.
The final frame-dependent landmark feature is then:
\begin{equation}
\tilde{l}_{t,j} = \phi\!\left([\,l_{t,j} \,\|\, p_j\,]\right),
\end{equation}
where $\phi$ is a learnable projection and $\|$ denotes feature concatenation.

\vspace{-3mm}
\paragraph{Landmark to mesh cross attention.}
These features are mapped to the shape of the mesh features with a learned cross attention layer between the landmark features and the mesh features. For each vertex $i$, we define a query from its mesh feature $f_i$ and keys and values from landmark features so that:
\begin{equation}
    q_{i} = W_Q f_i, \qquad
    k_{t, j} = W_K \tilde{l}_{t, j}, \qquad
    v_{t,j} = W_V \tilde{l}_{t, j} .
\end{equation}
Attention weights and landmark-conditioned features at vertex $i$ are given by:
\begin{equation}
    \alpha_{t, i, j} = \text{softmax}_j\left( \frac{q_i^T k_{t,j}}{\sqrt{d_c}} \right), \qquad c_{t,i} = \sum_{j=1}^N \alpha_{t, i, j} v_{t, j},
\end{equation}
where $d_c$ is the common attention dimension. We then form the fused vertex feature to augment the mesh feature field to obtain a new frame-dependent feature field $(\Tilde{f}_{t,i})$ such that $\Tilde{f}_{t,i} = (f_i, g_t, c_{t,i})$.




\vspace{-2mm}
\paragraph{Deformation decoding.}
We employ a DiffusionNet-based decoder (DN) to regress the dense deformation field. Let  $\tilde{F}_t = (\tilde{f}_{t,i})_{i=1}^{n} \in \mathbb{R}^{n \times (d + D + d_c)}$
be the augmented feature field. While the augmented features $\Tilde{f}_{t,i}$ already encode landmark-conditioned information, a purely local MLP would treat each vertex independently and ignore intrinsic geometric relations across the surface. 
To better exploit the mesh structure and ensure smooth, geometrically consistent deformations, we leverage an intrinsic surface network as a deformation decoder. Formally, we define a DiffusionNet decoder mapping the time dependent feature field to time dependent displacement fields as $D_{\mathrm{DN}}(M, \tilde{F}_t) = \Delta V_t \in \mathbb{R}^{n \times 3}$.



By operating intrinsically on the mesh, the DN decoder propagates information across neighboring vertices, leading to smoother and topology-aware deformation fields. The predicted mesh at frame $t$ is then given by $\hat{M}_t = (\hat{V}_t, F)$, where $\hat{V}_t = V + \Delta V_t$.
%
%
Finally, for each forward pass, STM iterates over all frames of the predicted landmark sequence $STM(M, (\Delta \hat{L}_t)_{t=1}^{T}) = (\hat{M}_t)_{t=1}^{T}$.




\subsection{Training Strategy}

Although ATS and STM operate in different representation spaces (landmark displacements and vertex displacements, respectively), both modules are trained using the same motion supervision principle. Let \(Y_t \in \mathbb{R}^{K \times 3}\) denote a generic motion representation at frame \(t\), where \(K=N\) landmarks for ATS and \(K=n\) mesh vertices for STM. Let \(\hat{Y}_t\) be the corresponding prediction. We define a unified motion loss:
\begin{equation}
\resizebox{\linewidth}{!}{$
\mathcal{L}_{\mathrm{motion}}
=
\underbrace{\frac{1}{TK} \sum_{t=1}^{T}
\| Y_t - \hat{Y}_t \|_2^2}_{\mathcal{L}_{\mathrm{pos}}}
+
\lambda_v
\underbrace{\frac{1}{(T-1)K} \sum_{t=1}^{T-1}
\| v(Y_t) - v(\hat{Y}_t) \|_2^2}_{\mathcal{L}_{\mathrm{vel}}}
+
\lambda_a
\underbrace{\frac{1}{(T-2)K} \sum_{t=1}^{T-2}
\| a(Y_t) - a(\hat{Y}_t) \|_2^2}_{\mathcal{L}_{\mathrm{acc}}}
$}.
\label{eq:unified_loss}
\end{equation}
Here, the first- and second-order temporal differences are defined as:
\begin{equation}
v(Y_t) = Y_{t+1} - Y_t, \qquad
a(Y_t) = Y_{t+2} - 2Y_{t+1} + Y_t.
\end{equation}
This formulation is instantiated for \textbf{ATS}, where \(Y_t = \Delta L_t\) corresponds to landmark displacements, and for \textbf{STM}, where \(Y_t = \Delta V_t\) corresponds to dense vertex displacements.
Despite sharing the same loss, the two modules are trained independently.
This decoupling reduces computational cost (STM training does not require backpropagation through ATS diffusion sampling) and enables precomputation of static mesh features. Furthermore, modular training provides finer control: ATS can be adapted to new speech-emotion datasets independently of mesh topology, while STM can be optimized for new mesh families without modifying the motion generator. At inference time, the landmark motion predicted by ATS is fed to STM to obtain the final animated mesh sequence \((\hat{M}_t)_{t=1}^{T}\).

\vspace{-0.3cm}
\section{Experiments}

In this section, we describe the experimental protocol adopted to evaluate our method against state-of-the-art approaches. 
First, we introduce the datasets and evaluation metrics in~\Cref{sec:dataset}, then present the quantitative and qualitative results in~\Cref{sec:quantitative}. More results are reported in the supplementary material.

\vspace{-0.3cm}
\subsection{Datasets and Metrics}\label{sec:dataset}

We train and evaluate our framework on multiple datasets to assess its robustness and generalization capabilities. 
Thanks to the modular design of our approach, the two stages of the model can be trained independently, enabling the use of different datasets for each component. 
This flexibility allows our framework to generalize across identities, topologies, and emotional settings. In our experiments, we used the following datasets:

\smallskip
\noindent
\textbf{\emph{1} EmoVOCA}~\cite{nocentini2025emovocaspeechdrivenemotional3d} provides speech-driven 3D facial sequences with emotional expressions paired with neutral (non-expressive) audio signals. The dataset spans 11 emotional categories: \textit{afraid}, \textit{ashamed}, \textit{angry}, \textit{disgust}, \textit{happy}, \textit{sad}, \textit{drunk}, \textit{ill}, \textit{suspicious}, \textit{pleased}, and \textit{upset}. In total, it contains 15.840 registered 3D mesh sequences. The dataset is split into training, validation, and test sets using 8 identities for training and 2 identities each for validation and testing. All meshes are in FLAME~\cite{FLAME:SiggraphAsia2017} topology, consisting of 5,023 vertices and 9,976 faces.

\noindent
\textbf{\emph{2} MEAD-EMOTE}~\cite{emote}, provides 3D facial reconstructions derived from the MEAD dataset~\cite{mead}. It includes seven emotional categories (\textit{fear}, \textit{anger}, \textit{disgust}, \textit{happiness}, \textit{surprise}, \textit{sadness}, and \textit{contempt}) recorded at three intensity levels, in addition to a neutral expression. The dataset contains 31,013 sequences from 47 actors, which are split into 36 for training, 5 for validation, and 6 for testing. All meshes are in FLAME~\cite{FLAME:SiggraphAsia2017} topology.

\noindent
\textbf{\emph{3} VOCAset}~\cite{VOCA2019} contains mesh sequences of 12 actors performing 40 speeches at 60\,fps in a neutral state. Sequences last between 3 and 5 seconds. All meshes are in FLAME~\cite{FLAME:SiggraphAsia2017} topology.

\noindent
\textbf{\emph{4} BIWI}~\cite{BIWI_2010} includes 14 subjects articulating 40 sentences each, recorded at 25\,fps in a neutral state. Each sentence lasts approximately 5 seconds, and all meshes share a fixed topology. Due to GPU memory constraints, we use a downsampled version denoted as BIWI$_6$, which contains 3,895 vertices and 7,539 faces.

\noindent
\textbf{\emph{5} Multiface}~\cite{wuu2022multiface} consists of 13 identities performing up to 50 speech sequences of approximately 4 seconds each, sampled at 30\,fps in a neutral state. The meshes have a fixed topology with 5,471 vertices and 10,837 faces.

\smallskip
For EmoVOCA, MEAD-EMOTE, and VOCAset, landmark correspondences are directly available thanks to the shared FLAME~\cite{FLAME:SiggraphAsia2017} topology. Since all meshes are registered to the same template, vertex-level correspondence across identities and sequences is inherently guaranteed, allowing us to consistently extract 3D landmarks from fixed vertex indices. In contrast, BIWI and Multiface do not follow the FLAME topology. Therefore, for these datasets, we manually selected a set of semantically consistent landmarks on the corresponding mesh templates to ensure comparable landmark supervision across datasets.

\smallskip
\noindent
\textbf{Metrics:} 
Following prior work~\cite{FaceDiffuser_Stan_MIG2023, nocentini2024fixedtopologiesunregisteredtraining, peng2023emotalk}, we use the following metrics:
\begin{itemize}
    \item \textbf{LVE (Lip Vertex Error):} Maximum $L_2$ error computed over vertices in the mouth region, capturing speech- and emotion-related lip dynamics.
    
    \item \textbf{MVE (Max Vertex Error):} Maximum $L_2$ distance between predicted and ground-truth mesh vertices, serving as a global reconstruction metric.
    
    \item \textbf{FDD (Upper-Face Dynamic Deviation):} Measures the discrepancy between the standard deviation of upper-face vertices in the generated sequence and the ground truth.
    
    \item \textbf{DTW (Dynamic Time Warping):} Evaluates similarity between predicted and ground-truth lip vertex trajectories while allowing non-linear temporal alignment.
    
    \item \textbf{DFD (Discrete Fréchet Distance):} Measures similarity between lip vertex trajectories while preserving temporal ordering.
    
    \item $\boldsymbol{\delta_M}$ \textbf{(Mean Squared Displacement Error):} Computes the error magnitude between displacement vectors of consecutive frames.
    
    \item $\boldsymbol{\delta_{Cd}}$ \textbf{(Cosine Displacement Error):} Computes cosine distance between displacement vectors of consecutive frames, catching directional consistency.
\end{itemize}

\subsection{Experimental Results}\label{sec:quantitative}
We evaluate \textbf{FreeTalk} against state-of-the-art speech-driven 3D facial animation methods. 
Our primary objective is not to optimize performance for a single benchmark, but rather to demonstrate that FreeTalk achieves competitive accuracy while offering substantially improved generalization across identities, mesh topologies, and emotional settings. Thanks to its modular design, the two components of FreeTalk (ATS and STM) can be trained independently and even on different datasets. 
This flexibility allows the model to leverage heterogeneous supervision signals and adapt to datasets with different characteristics. 
Importantly, our framework does not require predefined landmark correspondences: when such annotations are unavailable, manually selected landmarks (as done for BIWI$_6$ and Multiface) are sufficient to obtain strong performance. 
\begin{table*}[!ht]
\centering
\caption{\textbf{Quantitative comparison on the MEAD-EMOTE test set.} 
We compare FreeTalk with state-of-the-art speech-driven 3D facial animation methods trained on MEAD-EMOTE. 
Lower values indicate better performance. 
Numbered settings follow the dataset order defined in Sec.~\ref{sec:dataset}: 
\emph{(1)} EmoVOCA, \emph{(2)} MEAD-EMOTE, \emph{(3)} VOCAset, \emph{(4)} BIWI$_6$, \emph{(5)} Multiface. 
For FreeTalk, ``\emph{(a,b + c,d)}'' denotes that ATS is trained on the datasets before ``+'' and STM on those after it.}
\vspace{-1mm}
\resizebox{0.7\textwidth}{!}{
\begin{tabular}{@{} l@{\hspace{0.2cm}}ccccccc@{}}
\toprule
 & LVE$\downarrow$ & MVE$\downarrow$ & FDD$\downarrow$ & DTW$\downarrow$ & DFD$\downarrow$ & $\delta_M$$\downarrow$ & $\delta_{Cd}$$\downarrow$ \\ 
\midrule

Faceformer \emph{(2)}   & \textbf{12.7} & \underline{1.87} & 14.4 & 3.41 & 13.3 & \underline{3.11} & 0.98 \\
EmoFace  \emph{(2)}    & \underline{12.8} & \textbf{1.85} & 14.9 & 3.46 & 13.3 & \textbf{3.10} & 0.99 \\
\midrule 
FreeTalk \emph{(1,2,4,5 + 1,2,4,5)} &  16.4 & 2.29 & \textbf{3.35} & 2.65 & 11.4 & 4.61 & 0.83 \\
FreeTalk \emph{(1,2,4,5 + 2)}  &  16.6 & 2.31 & \underline{3.52} & 2.65 & 11.4 & 4.60 & 0.83 \\
FreeTalk \emph{(2 + 1,2,4,5)}  & 13.7 & 2.27 & 4.93 & \underline{2.61} & \underline{11.1} & 3.23 & \underline{0.82} \\
FreeTalk \emph{(2 + 2)}   & 13.8 & 2.27 & 5.12 & \textbf{2.60} & \textbf{11.0} & 3.25 & \textbf{0.81} \\
\bottomrule
\end{tabular}
}
\label{tab:quantitative_comparison}
\vspace{-0.5cm}
\end{table*}

\begin{figure}[!t]
    \centering
    \includegraphics[width=\linewidth]{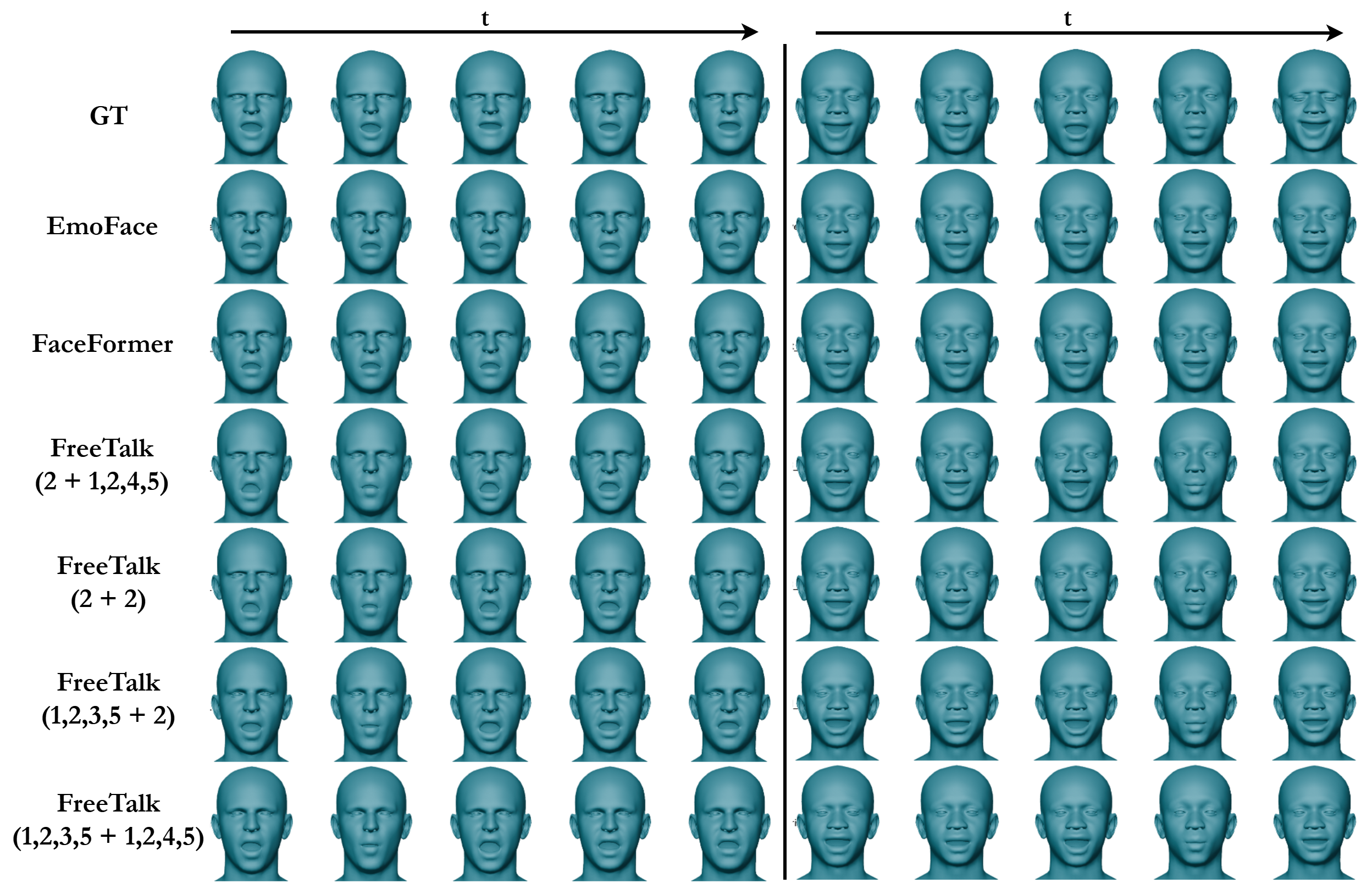}
    \caption{\textbf{Qualitative comparison on the MEAD-EMOTE test set.} Each panel shows a sequence of rendered frames for a single identity. \textbf{Left}: \textit{Disgust}; \textbf{right}: \textit{Happy}. Numbered settings follow the definition in                Table~\ref{tab:quantitative_comparison}.}
    \vspace{-0.7cm}
    \label{fig:quali}
\end{figure}

Table~\ref{tab:quantitative_comparison} reports quantitative results on the MEAD-EMOTE test set, comparing multiple training configurations of FreeTalk with state-of-the-art methods trained exclusively on MEAD-EMOTE. 
When both ATS and STM are trained solely on MEAD-EMOTE, FreeTalk achieves its best performance on this benchmark, reaching results comparable to specialized methods. 
Notably, even when trained on heterogeneous datasets, FreeTalk maintains competitive accuracy while gaining improved robustness to unseen identities and mesh topologies.
Figure~\ref{fig:quali} provides a qualitative comparison on two example sentences. 
Although some metrics in Table~\ref{tab:quantitative_comparison} slightly favor competing methods, the visual results show that FreeTalk produces more expressive and coherent emotional dynamics, while preserving accurate speech-related mouth movements. In particular, FreeTalk achieves the best performance in terms of FDD, a metric that evaluates upper-face dynamics, suggesting that our model generates more expressive facial movements compared to existing methods.

\begin{wrapfigure}{r}{0.5\textwidth}
    \centering
    \includegraphics[width=0.96\linewidth]{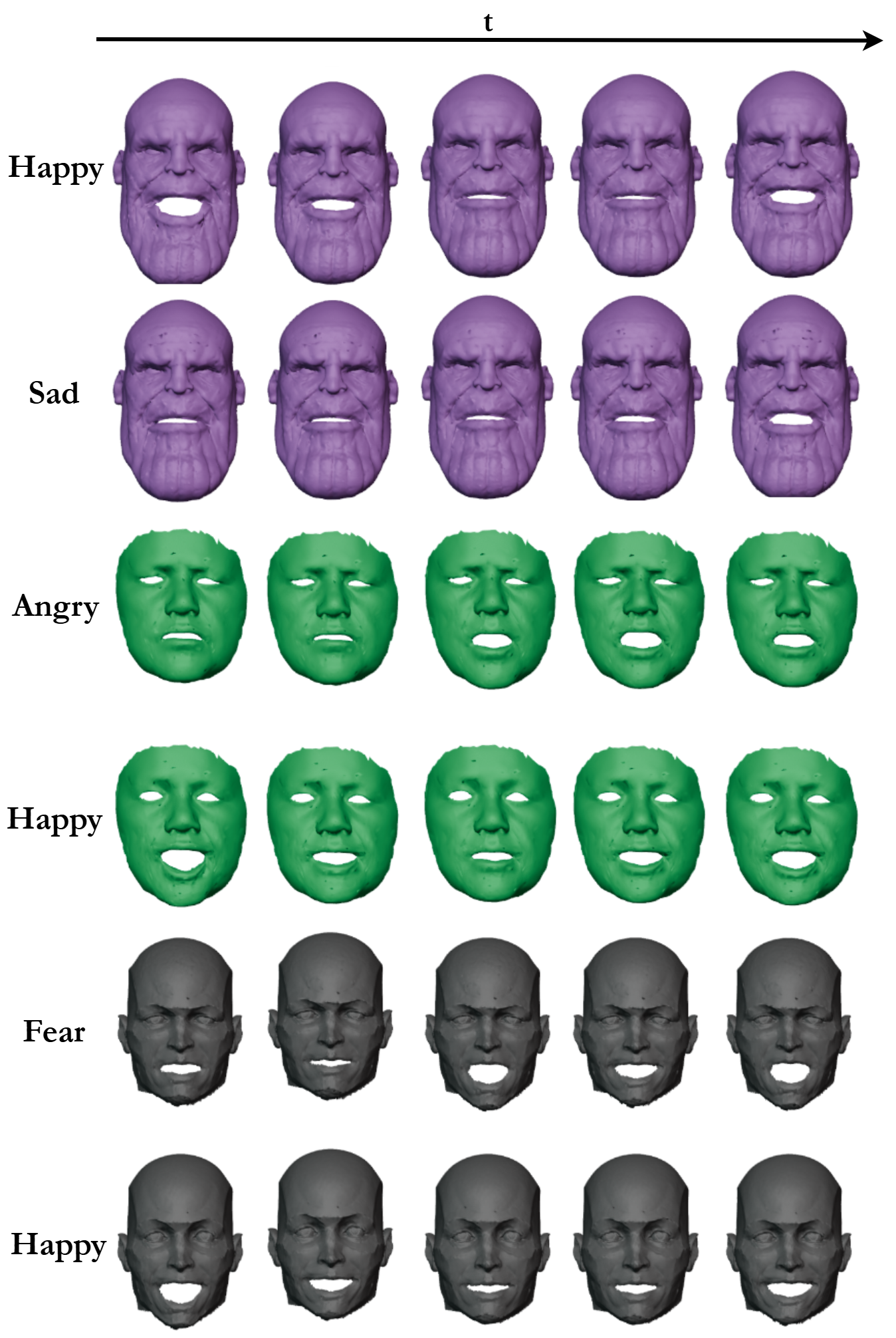}
    \vspace{-0.4cm}
    \caption{Results on unseen topologies.}
    \label{fig:scans}
\end{wrapfigure}

To demonstrate the generalization ability of FreeTalk,~\Cref{fig:scans} presents animations of identities unseen during training. The results highlight FreeTalk capability to generate expressive and coherent facial dynamics even on previously unseen mesh topologies, including highly challenging cases such as non-human or stylized characters where the general face structure is significantly different with respect to standard faces. 

In~\Cref{tab:unexpressive}, we evaluate both the performance and the generalization capability of FreeTalk. We compare several training variants of our model, obtained by using different combinations of datasets, against state-of-the-art methods on neutral (non-expressive) talking head benchmarks. The purpose of this analysis is not merely to surpass specialized methods, but to demonstrate that FreeTalk can achieve comparable performance while offering greater flexibility and cross-dataset robustness, thanks to the use of 3D landmarks as an intermediate representation. We study multiple configurations in which both submodules, ATS and STM, are trained on expressive, unexpressive, or mixed datasets. When training on unexpressive datasets, we set the emotion label to \emph{neutral} and the intensity to 1 to ensure consistency.

As expected, models trained exclusively on expressive data exhibit slightly degraded performance when evaluated on unexpressive test sets, since they learn facial dynamics that are not present in neutral sequences. In contrast, when FreeTalk is trained on unexpressive datasets, it achieves performance comparable to state-of-the-art methods specifically designed for neutral talking-head benchmarks. Notably, although FreeTalk is trained only in multi-dataset settings, it remains competitive with methods trained on a single dataset as well as with ScanTalk~\cite{nocentini2024scantalk3dtalkingheads}, which is evaluated in both single- and multi-dataset regimes. This suggests that the landmark-based intermediate representation enables the model to effectively integrate heterogeneous training data without sacrificing animation fidelity. Furthermore, the modular training of ATS and STM allows the framework to leverage different datasets with varying characteristics, improving robustness across identities, mesh topologies, and recording conditions. These results indicate that FreeTalk preserves animation quality across heterogeneous training conditions while additionally enabling controllable emotional dynamics in the generated meshes.
\vspace{-0.35cm}
\begin{table*}[!t]
\centering
\caption{\textbf{Cross-dataset generalization results on neutral speech datasets.} Lower values indicate better performance. 
\emph{(sd)} denotes single-dataset training. Numbered settings follow the definition in                Table~\ref{tab:quantitative_comparison}.}
\vspace{-0.2cm}
\label{tab:unexpressive}
\resizebox{\textwidth}{!}{
\begin{tabular}{@{}l@{}c@{\hspace{0.1cm}}c@{\hspace{0.1cm}}c@{\hspace{0.1cm}}c@{\hspace{0.1cm}}c@{\hspace{0.1cm}}c@{\hspace{0.1cm}}c@{\hspace{0.3cm}}|c@{\hspace{0.1cm}}c@{\hspace{0.1cm}}c@{\hspace{0.1cm}}c@{\hspace{0.1cm}}c@{\hspace{0.1cm}}c@{\hspace{0.1cm}}c@{\hspace{0.3cm}}|c@{\hspace{0.1cm}}c@{\hspace{0.1cm}}c@{\hspace{0.1cm}}c@{\hspace{0.1cm}}c@{\hspace{0.1cm}}c@{\hspace{0.1cm}}c@{}}
& \multicolumn{7}{c}{\textbf{\Large VOCAset}} 
& \multicolumn{7}{c}{\textbf{\Large BIWI\textsubscript{6}}} 
& \multicolumn{7}{c}{\textbf{\Large Multiface}} \\
\midrule
& LVE$\downarrow$ & MVE$\downarrow$ & FDD$\downarrow$ & DTW$\downarrow$ & DFD$\downarrow$ & $\delta_{M}\downarrow$ & $\delta_{Cd}\downarrow$ 
& LVE$\downarrow$ & MVE$\downarrow$ & FDD$\downarrow$ & DTW$\downarrow$ & DFD$\downarrow$ & $\delta_{M}\downarrow$ & $\delta_{Cd}\downarrow$ 
& LVE$\downarrow$ & MVE$\downarrow$ & FDD$\downarrow$ & DTW$\downarrow$ & DFD$\downarrow$ & $\delta_{M}\downarrow$ & $\delta_{Cd}\downarrow$ \\
\midrule
VOCA \emph{(sd)} & 6.99 & 0.98 & 2.66 & 1.77 & 7.41 & 1.39 & 1.02 
     & 5.74 & 2.59 & 41.5 & 1.60 & 7.89 & 1.47 & 0.67 
     & 4.92 & 2.76 & 55.78 & 1.56 & 6.51 & 0.86 & 1.23 \\
FaceFormer \emph{(sd)} & 6.12 & 0.93 & 2.16 & 1.33 & 5.39 & 0.86 & 0.58 
           & 4.08 & 2.16 & 37.1 & 1.56 & 6.61 & \underline{0.87} & \textbf{0.55} 
           & 2.45 & \textbf{1.45} & 20.2 & \textbf{0.87} & \underline{4.13} & \textbf{0.22} & \textbf{0.73} \\
FaceDiffuser \emph{(sd)} & 4.35 & 0.90 & 2.43 & 1.73 & 6.83 & 0.81 & 0.60 
             & \underline{4.02} & 2.12 & 39.6 & 1.55 & \textbf{6.50} & \textbf{0.85} & \underline{0.56} 
             & 3.55 & 2.39 & 29.1 & 1.45 & 5.24 & \underline{0.28} & 0.77 \\
CodeTalker \emph{(sd)} & 3.55 & \underline{0.89} & 2.26 & 1.33 & 5.66 & 0.80 & \underline{0.55} 
           & 5.19 & 2.64 & \textbf{20.6} & \underline{1.50} & 6.61 & 0.91 & 0.59 
           & 4.09 & 2.38 & 47.9 & 1.44 & 5.95 & 0.55 & 0.96 \\
SelfTalk \emph{(sd)} & 5.61 & 0.91 & 2.32 & 1.25 & 5.43 & 0.72 & 0.56 
         & \textbf{3.63} & \textbf{2.06} & 35.5 & 1.60 & 6.93 & 1.03 & 0.66 
         & \textbf{2.28} & 1.90 & 37.4 & 0.96 & \underline{4.18} & 0.25 & 0.75 \\
ScanTalk \emph{(sd)} & \textbf{3.01} & \textbf{0.86} & 2.40 & \underline{1.20} & 5.25 & 0.73 & \underline{0.55}  
                    & 4.65 & 2.14 & 36.0 & 1.56 & 7.22 & 1.05 & 0.63 
                    & 2.65 & 1.87 & 64.4 & 0.96 & 4.27 & 0.57 & 0.93 \\
ScanTalk \emph{(3,4,5)} & 7.05 & 1.07 & \textbf{1.29} & \textbf{1.19} & 5.25 & 0.72 & \textbf{0.52} 
                    & 4.44 & \underline{2.09} & 36.6 & \textbf{1.46} & 6.71 & 1.01 & 0.61 
                    & \underline{2.37} & \underline{1.71} & 16.0 & \underline{0.93} & 4.71 & 0.33 & \underline{0.74} \\
\midrule
FreeTalk \emph{(1,2,4,5 + 1,2,4,5)} & 4.28 & 0.99 & \underline{1.38} & 1.87 & 5.87 & 0.39 & 0.73  
                       & 6.10 & 2.45 & 40.3 & 1.55 & 6.84 & 1.82 & 0.76  
                       & 10.7 & 2.39 & \underline{5.34} & 1.81 & 6.89 & 0.71 & 0.90 \\
FreeTalk \emph{(1,2,4,5 + 3,4,5)} & 4.01 & 1.01 & 1.67 & 1.73 & 5.69 & 0.36 & 0.74  
                       & 5.58 & 2.37 & 30.8 & 1.56 & 7.14 & 1.76 & 0.75  
                       & 7.89 & 2.31 & \textbf{4.01} & 1.58 & 6.35 & 0.65 & 0.89 \\
FreeTalk \emph{(3,4,5 + 1,2,4,5)} & 3.12 & 0.94 & 1.56 & 1.61 & \textbf{5.08} & \underline{0.33} & 0.67  
                       & 4.72 & 2.20 & 36.0 & 1.52 & \underline{6.54} & 1.13 & 0.63  
                       & 3.90 & 1.95 & 16.4 & 1.19 & 5.36 & 0.44 & 0.77 \\
FreeTalk \emph{(3,4,5 + 3,4,5)} & \underline{3.05} & 0.95 & 1.85 & 1.60 & \underline{5.16} & \textbf{0.31} & 0.69  
                       & 4.44 & 2.17 & \underline{30.0} & 1.53 & 6.87 & 1.11 & 0.63  
                       & 3.19 & 1.92 & 14.3 & 1.14 & 5.29 & 0.46 & 0.76 \\
\bottomrule
\end{tabular}}
\end{table*}

\section{Conclusions}
We presented \textbf{FreeTalk}, a two-stage framework for emotionally expressive, speech-driven 3D facial animation that generalizes across \emph{arbitrary} mesh topologies without template registration. FreeTalk first synthesizes emotion-conditioned facial motion as landmark displacement sequences via ATS, and then transfers this sparse motion to dense mesh deformations on any target geometry via STM using intrinsic surface learning. Experiments demonstrate that our approach achieves competitive accuracy with state-of-the-art methods on expressive benchmarks while substantially improving generalization across unseen mesh topologies and even non-human or stylized faces, inaccessible to template-based methods. 

Still, FreeTalk exhibits limitations. First, ATS treats emotions as a discrete categorical label (with intensity), while a continuous latent emotion space could enable finer-grained stylization. 
Second, STM uses DiffusionNet, requiring to solve the eigenvalue problem of the Laplace-Beltrami operator. This operation can be costly for large meshes. PoissonNet~\cite{maesumi2025poissonnet} addresses this issue but showed slightly worse accuracy (see ablation in the supp. material). Additionally, the Laplace-Beltrami operator is intrinsic to the mesh and can lose robustness when large tears and holes are present.

\section*{Acknowledgments}
This work is supported by the    \href{https://geogen3dhuman.univ-lille.fr}{CNRS Int. Research Project GeoGen3DHuman}. This work was also partially supported by ``Partenariato FAIR (Future Artificial Intelligence Research) - PE00000013, CUP J33C22002830006" funded by NextGenerationEU through the italian MUR within NRRP, project DL-MIG. This work was also partially funded by the ministerial decree n.352 of the 9th April 2022 NextGenerationEU through the italian MUR within NRRP. This work was also partially supported by the project 4DSHAPE ANR-24-CE23-5907 of the French National Research Agency (ANR), and by Fédération de Recherche Mathématique des Hauts-de-France (FMHF, FR2037 du CNRS). 
This work was also partially supported by the AI4Debunk project (HORIZON-CL4-2023-HUMAN-01-CNECT grant n.101135757).

\title{Supplementary Material}

\author{Federico Nocentini$^*$\inst{1}
\and Thomas Besnier$^*$\inst{2}
\and  Claudio Ferrari\inst{1}
\and  Stefano Berretti\inst{1}
\and Mohamed Daoudi\inst{3}}
\authorrunning{F. Nocentini et al.}
%

\institute{
 Media Integration and Communication Center (MICC),\\ University of Florence, Italy\\ 
 \email{federico.nocentini@unifi.it, stefano.berretti@unifi.it, claudio.ferrari@unifi.it}
\and University of Copenhagen
 \email{thomas.besnier@di.ku.dk} \\ \and
IMT Nord Europe, Institut Mines-Télécom, Centre for Digital Systems
\email{mohamed.daoudi@imt-nord-europe.fr}\\
}

\maketitle

\section{Ethical statement}
FreeTalk is thought as beneficial for applications in virtual reality, digital human modeling and accessibility tools. However, as for any technology capable of synthesizing realistic face motions, we acknowledge the risk of potential misuse. In particular, we strongly condemn impersonation, deceptive synthetic media or any application that animates the representation of any individual without their informed and explicit consent. While FreeTalk operates on 3D meshes, it requires a 3D scan of a target individual as input, which raises the practical barrier for malicious usage compared to 2D videos but the risk is still present.\\
Regarding data, all experiments rely on publicly available datasets collected under institutional oversight with participant consent; no new human subject data was collected, and no personally identifiable information beyond what is present in the original published datasets is used or disclosed.\\
After the reviewing process, we commit to releasing code and pretrained weights under a license that explicitly prohibits non-consensual animation of real individuals.

\section{Implementation details}
\paragraph{Hardware.}
All experiments were conducted on a workstation equipped with a single NVIDIA GeForce RTX 5090 GPU with 32\,GB of VRAM. Training was performed using mixed precision to improve computational efficiency.

\paragraph{Hyperparameters.}
FreeTalk is composed of two modules trained independently: an audio-to-landmark generator and a landmark-to-mesh densifier (STM). 
The audio-to-landmark generator is implemented as a transformer-based diffusion model with hidden dimension $d_{model}=512$, $6$ transformer layers, $8$ attention heads, and a feed-forward dimension of $2048$. The diffusion process uses $1000$ training timesteps and $100$ DDIM sampling steps. Models are trained for $100$ epochs using the AdamW optimizer with learning rate $10^{-4}$ and weight decay $10^{-2}$, batch size $16$, and gradient clipping set to $1.0$. Temporal smoothness is encouraged using velocity and acceleration losses weighted by $0.3$ and $0.1$, respectively.

The STM module uses DiffusionNet as both encoder and decoder with feature width $128$ and latent dimension $128$. Landmark conditioning is performed through a graph convolutional network with $3$ layers and hidden dimension $128$, followed by a cross-attention module with $4$ heads and model dimension $128$. The STM is trained for $30$ epochs using AdamW with learning rate $10^{-4}$ and weight decay $10^{-2}$, and employs velocity and acceleration losses weighted by $0.5$ and $0.2$ to improve temporal consistency.

\section{Ablation / Component substitution study for STM}\label{sec:ablation}


This module predicts a mapping from landmark displacements to a dense displacement field on a face mesh of any arbitrary topology. In particular, it is made of a mesh feature extractor, a feature combination and a point-wise deformation decoder. Compared to the main paper, the training and testing datasets are subsamples of the full datasets: all sequences are sampled at 10 fps (between 15 and 30 frames per sequence) and only half of the sequences are used. This enables us to run ablation and component substitution studies much quicker.

\noindent\textbf{Static mesh feature extractor and decoder.} First, we report a component substitution study regarding the mesh feature extractor and the deformation decoder. In~\Cref{tab:component_substitution}, we compare the Lip Vertex Error (LVE) in different configurations to account for the accuracy of predicted deformation sequences.

\begin{table}[ht!]
\centering
\caption{\textbf{Component substitution analysis for the STM module.} We report the Lip Vertex Error (LVE) of the STM module when trained and tested on VOCAset, BIWI and MultiFace. Each cell correspond to a different model for the mesh feature extractor (left column) and the deformation decoder (first row).}
\begin{tabular}{l@{\hspace{1cm}}c@{\hspace{0.5cm}}c@{\hspace{0.5cm}}c@{\hspace{0.5cm}}}
\toprule
\multirow{2}{*}{Feature extractor} & \multicolumn{3}{c}{Deformation Decoder} \\ \cline{2-4} 
              & Shared MLP      & DiffusionNet      & NJF~\cite{NeuralJacobianField_2022}      \\ \hline
Shared MLP   & 0.0020 & 0.0023 &  0.0018      \\ 
DiffusionNet~\cite{SharpDiffusionNet} & 0.0017    &      0.0021         &     0.0018     \\
PoissonNet~\cite{maesumi2025poissonnet}    &   0.0016   &   0.0020   & 0.0017  \\ \bottomrule
\end{tabular}
\label{tab:component_substitution}
\end{table}

\begin{figure}[!th]
    \centering
    \includegraphics[width=0.8\linewidth]{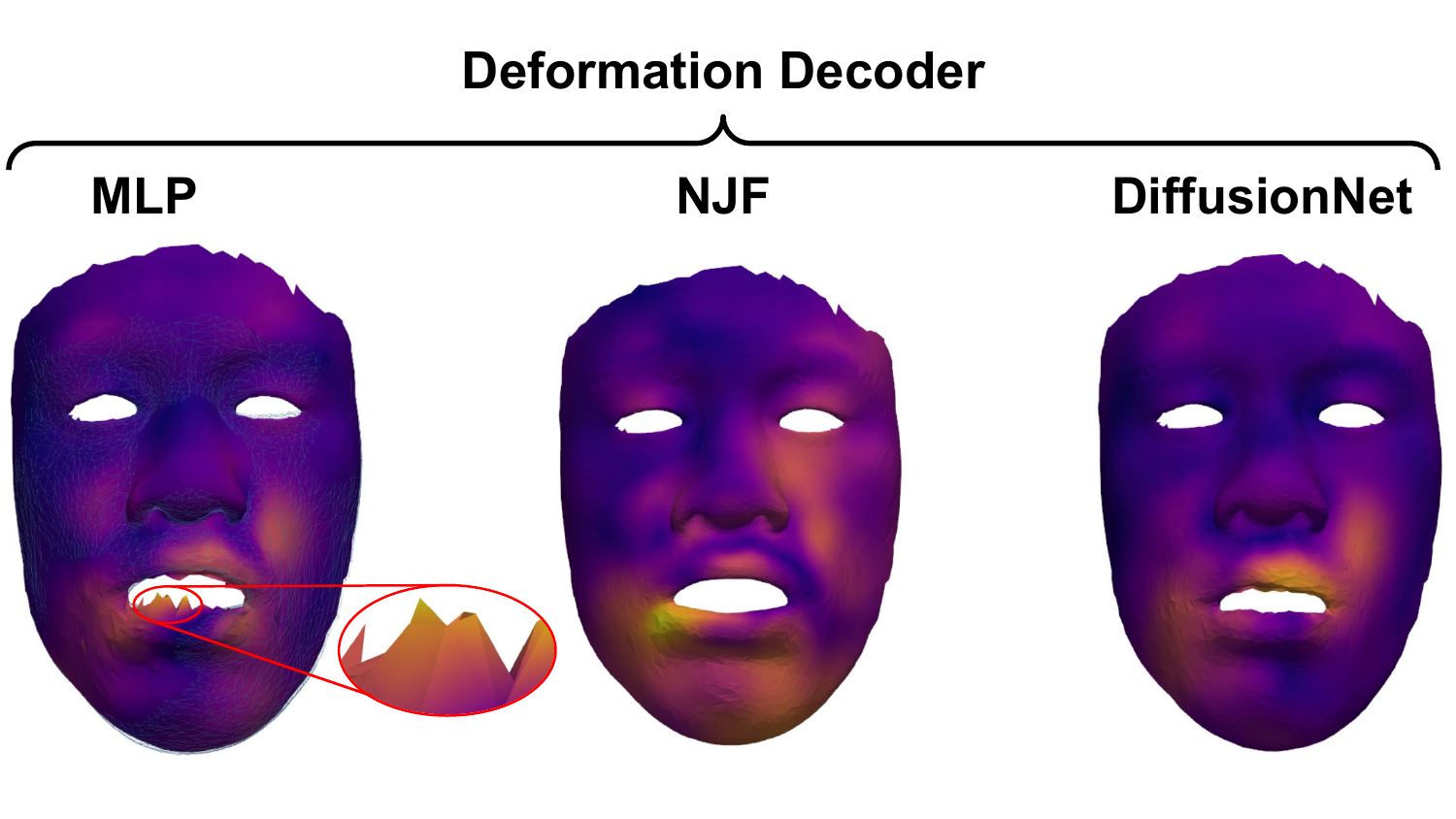}
    \vspace{-0.5cm}
    \caption{\textbf{Vertex-wise L2 error on a predicted deformed mesh.} We use different models for the deformation decoder. We highlight how a MLP as decoder tends to predict noisy deformations around the lips. On the opposite, Neural Jacobian Fields tends to smooth out "too much" to capture highly localized deformations around the right of the lower lip area.}
    \label{fig:supp_heatmap_ablation}
\end{figure}

In this table, we observe that reported performance with respect to the LVE metric is mostly comparable across the different Feature extractor/deformation decoder configurations. However, qualitatively, with a MLP as deformation decoder, predicted deformations are "noisy" as shown by~\Cref{fig:supp_heatmap_ablation}. On the opposite, Neural Jacobian Fields (NJF) \cite{NeuralJacobianField_2022} have demonstrated to produce smooth deformations but then, predicted deformation fields are globally smoothed by solving the Poisson equation on the mesh, which is inherently global and offer less locality than propagating geometric features with the heat equation. Because of this, using DiffusionNet\cite{SharpDiffusionNet} for the deformation decoder yields more satisfying results than a MLP or Neural Jacobian Fields despite a slightly worse LVE.

\noindent\textbf{Feature combination. }Next, we support the design choice for the feature combination strategy to fuse the sparse features from the landmark displacements to the static mesh features. For this, we report in~\Cref{tab:ablation_feat_combination} the LVE to account for raw accuracy of the predicted deformation sequences.

\begin{table}[ht!]
    \centering
        \caption{Ablation studies for the feature combination component of the STM module.}
    \begin{tabular}{l@{\hspace{0.5cm}}c}
    \toprule
    Feature combination strategy & LVE   \\ \hline
    Concat.     &   0.0035  \\
    CA          &   0.0025     \\
    CA + Concat.     &  0.0023  \\    
    GCN + CA + Concat. (ours) &  0.0021  \\ \bottomrule 
    \end{tabular}
\label{tab:ablation_feat_combination}
\end{table}
In particular, we observe a large drop in performance when we only use raw displacement vectors. This makes sense as a given displacement vector can be attributed to different positions on the face during the deformation sequence. To address this, the cross-attention between landmark displacements (with positional embedding based on the graph node index) really allow the model to learn that a displacement vector on a given node of the landmark graph should be transferred to a specific set of vertices on the dense unregistered mesh. Adding graph convolutional features further improves the performance by linking neighboring displacement vectors on the landmark graph.

\begin{figure}[!th]
    \centering
    \includegraphics[width=1.0\linewidth]{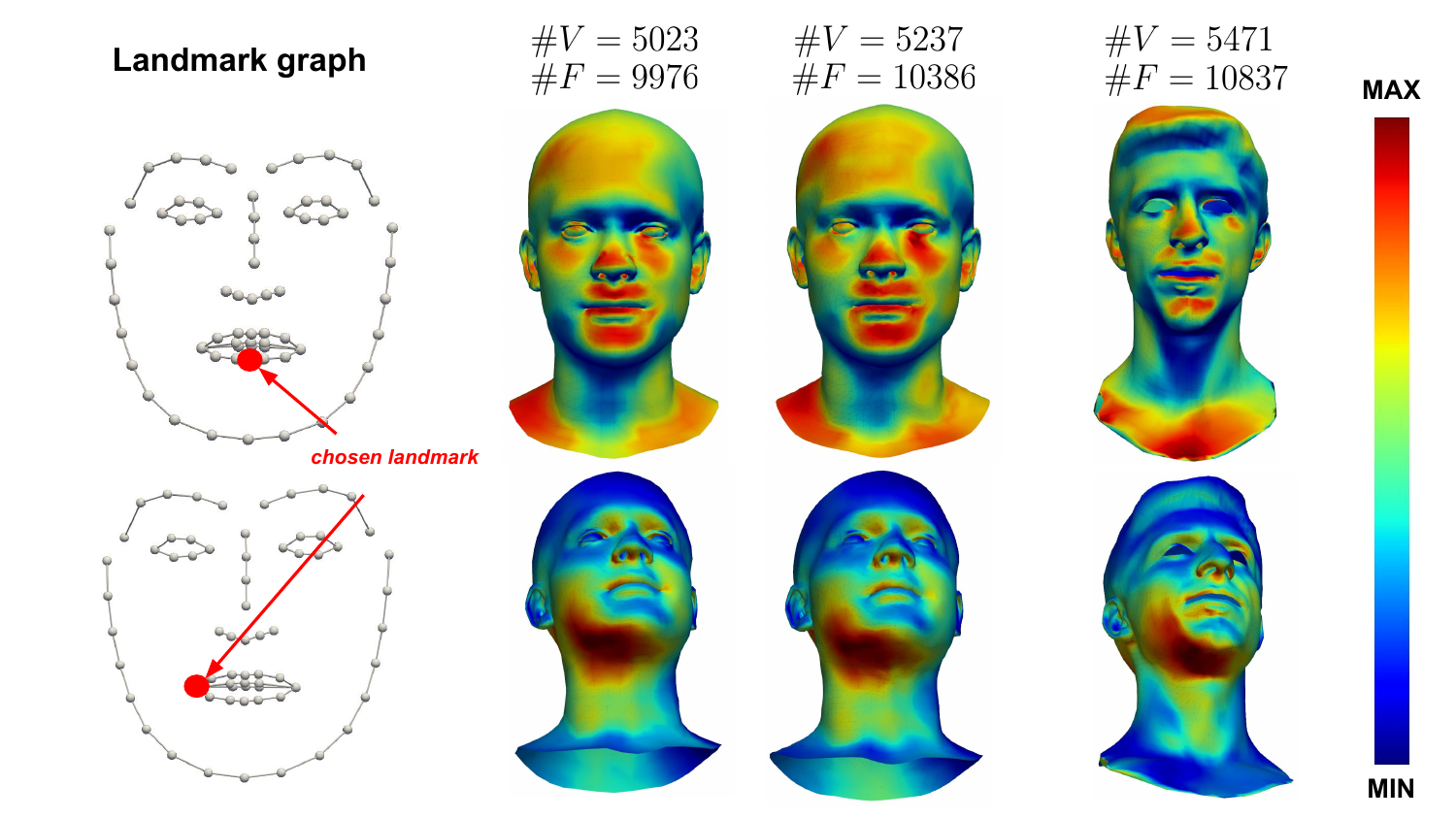}
    \caption{\textbf{Heatmap of the attention scores from one landmark to each mesh vertex at one frame.} For a selected frame and landmark (in \textcolor{red}{red} on the landmark graph on the left of the figure), how strongly each mesh vertex attends to that landmark inside the model’s cross-attention module. Red values mean that the vertex relies more on the selected landmark when building its time-dependent latent feature, while blue values mean that the landmark has less influence there.}
    \label{fig:attention_viz}
\end{figure}

\textbf{Qualitative view of the attention map from landmark displacements to dense unregistered meshes.} One of the technical contribution of FreeTalk comes from the learned cross attention between landmark features and vertex features. Basically, the intuitive thing here is that, for a fixed frame, each vertex has to find the "most relevant landmarks" to drive its deformation. We highlight this with a heatmap in~\Cref{fig:attention_viz}. It shows, for a selected frame and landmark (in \textcolor{red}{red} on the landmark graph on the left of the figure), how strongly each mesh vertex attends to that landmark inside the model’s cross-attention module. Red values mean that the vertex relies more on the selected landmark when building its time-dependent latent feature, while blue values mean that the landmark has less influence there.

These values should be interpreted as relative attention weights, not as geometric distances, displacement magnitudes, or explicit correspondences between landmarks and mesh vertices. Since the mesh is unregistered, the heatmap indicates which regions of the mesh are most affected by the selected landmark in the learned feature space, rather than where that landmark is physically located on the mesh. In particular, we highlight how robust the cross attention is to remeshing and identity change thanks to the robustness of the underlying model.

\bibliographystyle{splncs04}
\bibliography{main}
\end{document}